# Heat Diffusion Models - Interpixel Attention Mechanism

[1]Pengfei Zhang, [2]Shouqing Jia

*[1,2]Northeastern University at Qinhuangdao*
*Email: [2]jiashouqing@163.com*

**Abstract**

Denoising Diffusion Probabilistic Models (DDPM) process images as a whole. Since adjacent pixels are highly likely to belong to the same object, we propose the Heat Diffusion Model (HDM) to further preserve image details and generate more realistic images. HDM essentially is a DDPM that incorporates an attention mechanism between pixels. In HDM, the discrete form of the two-dimensional heat equation is integrated into the diffusion and generation formulas of DDPM, enabling the model to compute relationships between neighboring pixels during image processing. Our experiments demonstrate that HDM can generate higher-quality samples compared to models such as DDPM, Consistency Diffusion Models (CDM), Latent Diffusion Models (LDM), and Vector Quantized Generative Adversarial Networks (VQGAN).

## 1. INTRODUCTION

Generative models have demonstrated their capability to produce high-quality samples across numerous domains. Variational Autoencoders (VAE), and Generative Adversarial Networks (GAN) are increasingly fading from the spotlight[1]. Diffusion-based image generation models have emerged as one of the preferred approaches for addressing challenges in the field of image generation[2]. For example, Denoising Diffusion Probabilistic Models (DDPM) introduces a stepwise denoising process on the basis of DM, significantly improving the quality of generated images and the stability of training[3]. The CDM aims to enhance generation efficiency by reducing sampling steps while maintaining high-quality image synthesis. By incorporating consistency constraints, CDM optimizes the training process, mitigating issues such as mode collapse or instability that are often encountered in traditional diffusion models. This makes CDM particularly advantageous in scenarios requiring real-time performance[4]. In addition to diffusion models, approaches such as the Masked Generative Image Transformer (MaskGIT) utilize a masked prediction mechanism to generate images by progressively predicting masked regions of the image[5]. This method reduces computational complexity during the generation process while improving generation efficiency[6]. Certainly, the Visual AutoRegressive (VAR) model is also a viable option. It integrates the Transformer architecture and enhances generation efficiency and quality through "next-scale prediction"[7]. In the field of AR, methods such as the Vector Quantized Generative Adversarial Network (VQGAN), which integrates VQ-VAE and GAN, the Vision Transformer with Vector Quantization (ViTVQ) based on Vision Transformer (ViT)[8], and the large-scale pre-trained LlamaGen-B have significantly enriched the domain of image generation.

However, in the process of progressively generating high-resolution images from low-resolution inputs, the VAR model may exhibit the issue where prediction errors at early scales can be amplified, leading to inconsistencies or distortions in local details of the final image[9]. Although MaskGIT performs well in capturing global structures, local details are prone to distortion when generating high-resolution images. Models such as VQGAN and ViTVQ also suffer from the issue of detail blurring when generating complex scenes[10]. And a limitation of DDPMs and CDM lies in their uniform processing of all pixels during noise addition or removal, without assigning specialized weights to semantically critical regions[11]. This results in suboptimal preservation of fine-grained details[12]. In recent years, the attention mechanism has shown great potential, so we are considering incorporating pixel-level attention into the DDPM[13]. Embedding a pixel-wise interaction mechanism (for instance, gradient propagation governed by the principles of thermal diffusion) within the diffusion process enables dynamic coordination of signal evolution within local regions at each denoising step[14], thereby achieving better alignment with the inherent continuity priors of natural images[15]. Such refinement allows the generation process to more accurately capture coherent micro textures (e.g. hair) while mitigating edge blurring or localized artifacts caused by isolated pixel optimization[16]—all achieved without compromising compatibility with the original DDPM training paradigm or introducing additional adversarial training components or architectural complexities.

Section 2 provides a detailed description of the proposed HDM and its modifications to DDPM. Section 3 presents experimental results comparing HDM, DDPM, and other models under identical test datasets, and Section 4 gives the conclusion.

## 2. HEAT DIFFUSION MODELS

In this paper, to incorporate the relationships between adjacent pixels into the computation process of DDPM, the discretized heat equation is selected, and the rationale for choosing the heat equation will be explained step-by-step in the subsequent formula derivations.

Given $u = u(x, y, \mathrm{t})$, it expresses the temperature at position $(x, y)$ and time $t$. The heat equation is:

$$\frac{\partial u}{\partial t} - \kappa \nabla^2 u = 0 \tag{1}$$

where $\kappa$ denotes the thermal diffusion coefficient and $\nabla^2 = \frac{\partial^2}{\partial x^2} + \frac{\partial^2}{\partial y^2}$ represents the spatial Laplacian operator. The physical essence of the equation lies in its description of energy equilibrium driven by local gradients, i.e., heat diffuses from regions of high gradient to those of low gradient. This is highly similar to the gradients employed in score-based generative models, where gradient-guided learning drives the generated samples to asymptotically approach the true data distribution. Take $x = i\Delta$, $y = j\Delta$, $t = n\tau$, the result after discretization is:

$$u_n^{i,j} = u(i\Delta, j\Delta, n\tau)$$

Digital images are essentially two-dimensional discrete grids, which are completely isomorphic to the difference grids used in the numerical solution of the heat equation. Therefore, we can regard $i$ and $j$ as the position coordinates of a certain pixel in the image. Performing first-order difference approximation and second-order difference approximation, the following forms are obtained:

$$\frac{\partial u}{\partial t}\Big|_n^{i,j} \approx \frac{u_{n+1}^{i,j} - u_n^{i,j}}{\tau} \quad \text{(First-order forward difference)}$$

$$\frac{\partial u}{\partial t}\Big|_n^{i,j} \approx \frac{u_n^{i,j} - u_{n-1}^{i,j}}{\tau} \quad \text{(First-order backward difference)}$$

$$\frac{\partial u}{\partial x}\Big|_n^{i+\frac{1}{2},j} \approx \frac{u_n^{i+1,j} - u_n^{i,j}}{\Delta} \quad \text{(Central difference at } i + \tfrac{1}{2})$$

$$\frac{\partial^2 u}{\partial x^2}\Big|_n^{i,j} \approx \frac{u_n^{i+1,j} - 2u_n^{i,j} + u_n^{i-1,j}}{\Delta^2} \quad \text{(Second-order difference)}$$

Applying forward difference to the temporal of the heat equation (1) at $t = n\tau$ and backward difference to the temporal of the heat equation (1) at $t = (n+1)\tau$ yields:

$$u_{n+1}^{i,j} - u_n^{i,j} = K\left(u_n^{i+1,j} + u_n^{i-1,j} + u_n^{i,j+1} + u_n^{i,j-1} - 4u_n^{i,j}\right)$$

$$u_{n+1}^{i,j} - u_n^{i,j} = K\left(u_{n+1}^{i+1,j} + u_{n+1}^{i-1,j} + u_{n+1}^{i,j+1} + u_{n+1}^{i,j-1} - 4u_{n+1}^{i,j}\right)$$

where:

$$K = \frac{\kappa\tau}{\Delta^2}$$

A weighted summation of the above two equations yields (where $0 \le \theta \le 1$):

$$u_{n+1}^{i,j} - u_n^{i,j} = \theta K\left(u_n^{i+1,j} + u_n^{i-1,j} + u_n^{i,j+1} + u_n^{i,j-1} - 4u_n^{i,j}\right)$$

$$+(1-\theta)K\left(u_{n+1}^{i+1,j} + u_{n+1}^{i-1,j} + u_{n+1}^{i,j+1} + u_{n+1}^{i,j-1} - 4u_{n+1}^{i,j}\right)$$

The scheme reduces to fully explicit when $\theta = 1$, fully implicit when $\theta = 0$, and Crank-Nicolson when $\theta = 1/2$[18]. The above equation can be rearranged as(valid for $0 < i < I - 1,\ 0 < j < J - 1$):

$$(1 + 4(1-\theta)K)u_{n+1}^{i,j} - (1-\theta)K\left(u_{n+1}^{i+1,j} + u_{n+1}^{i-1,j} + u_{n+1}^{i,j+1} + u_{n+1}^{i,j-1}\right)$$

$$= (1 - 4\theta K)u_n^{i,j} + \theta K\left(u_n^{i+1,j} + u_n^{i-1,j} + u_n^{i,j+1} + u_n^{i,j-1}\right) \quad (2)$$

However, the given equation fails to determine the values at the positions $i = 0, I - 1$, $j = 0, J - 1$, and its adverse effects cannot be predicted a priori; therefore, an appropriate boundary condition should be selected. A common boundary condition involves specifying fixed values at the boundaries. Here, the boundary conditions are assumed to be 0.

To avoid affecting the subsequent reasoning process due to the complexity of the formulas, we introduce the matrix form, arranging $u_n^{i,j}$ as a column vector($0 \leq i \leq I-1,\ 0 \leq j \leq J-1$), for ease of calculation, the values of $I$ and $J$ here directly represent the number of rows and columns of the image dimensions, and obtain:

$$\boldsymbol{u}_n = \begin{bmatrix} u_n^{0,0} \\ \vdots \\ u_n^{0,J-1} \\ \vdots \\ u_n^{I-1,0} \\ \vdots \\ u_n^{I-1,J-1} \end{bmatrix}$$

At this point, Eq. (2) is written in matrix form:

$$\boldsymbol{S}\boldsymbol{u}_{n+1} = \boldsymbol{T}\boldsymbol{u}_n$$

The boundary condition used here is adiabatic condition:

$$\left.\frac{\partial u}{\partial x}\right|_n^{0,j} = \left.\frac{\partial u}{\partial x}\right|_n^{I-1,j} = \left.\frac{\partial u}{\partial y}\right|_n^{i,0} = \left.\frac{\partial u}{\partial y}\right|_n^{i,J-1} = 0$$

Taking the boundary at $i = 0,\ 1 < j < J-1$ as an example, the discrete form of $\left.\frac{\partial u}{\partial x}\right|_n^{0,j} = 0$ is obtained:

$$\frac{u_n^{1,j} - u_n^{-1,j}}{2\Delta} = 0$$

This yields the relation $u_n^{-1,j} = u_n^{1,j}$ ,which when substituted into Eq. (2) leads to the following form:

$$(1 + 4(1-\theta)K)u_{n+1}^{0,j} - (1-\theta)K\left(2u_n^{1,j} + u_{n+1}^{0,j+1} + u_{n+1}^{0,j-1}\right)$$

$$= (1 - 4\theta K)u_n^{0,j} + \theta K\left(2u_n^{1,j} + u_n^{0,j+1} + u_n^{0,j-1}\right)$$

To better illustrate the form of the boundary conditions, the matrix **S** and **T** corresponding to $I = J = 2$ is explicitly expressed as follows:

$$\boldsymbol{S} = \begin{bmatrix} 1+4(1-\theta)K & 2(\theta-1)K & 0 & 2(\theta-1)K \\ 2(\theta-1)K & 1+4(1-\theta)K & 2(\theta-1)K & 0 \\ 0 & 2(\theta-1)K & 1+4(1-\theta)K & 2(\theta-1)K \\ 2(\theta-1)K & 0 & 2(\theta-1)K & 1+4(1-\theta)K \end{bmatrix}$$

$$\boldsymbol{T} = \begin{bmatrix} 1-4\theta K & 2\theta K & 0 & 2\theta K \\ 2\theta K & 1-4\theta K & 2\theta K & 0 \\ 0 & 2\theta K & 1-4\theta K & 2\theta K \\ 2\theta K & 0 & 2\theta K & 1-4\theta K \end{bmatrix}$$

From $\boldsymbol{S}\boldsymbol{u}_{n+1} = \boldsymbol{T}\boldsymbol{u}_n$, we can define $\boldsymbol{A} = \boldsymbol{S}^{-1}\boldsymbol{T}$, thereby achieving:

$$\boldsymbol{u}_{n+1} = \boldsymbol{A}\boldsymbol{u}_n$$

Once the matrix representing the positional and quantitative relationships between pixels is obtained, modifications to the diffusion process formula of DDPM can be initiated.

Initially, the diffusion formula can be expressed as:

$$\boldsymbol{u}_n = \sqrt{\alpha_n} A \boldsymbol{u}_{n-1} + \sqrt{1-\alpha_n}\boldsymbol{\varepsilon}_{n-1},\ \boldsymbol{\varepsilon}_n \sim \mathcal{N}(0, \boldsymbol{I})\ ,\ n = 0 \sim N$$

at this stage, matrix $\boldsymbol{A}$ is simply embedded into the formula, and in the subsequent derivation process, matrix $\boldsymbol{A}$ will impose substantial computational complexity. Not only that, but the appearance of $\boldsymbol{A}^n$ in subsequent calculations will result in the variance of the entire distribution no longer being equal to 1[17]. Maintaining a total variance of 1 ensures that the noise addition in the diffusion process remains controllable, preventing instability or convergence difficulties in the training process caused by excessively large or small variances[19]. A variance of 1 provides a standardized noise scale, enabling the model to learn the data distribution more stably[20]. And the natural images typically exhibit local smoothness and continuity, and the noise-adding process with a variance of 1 can better align with such prior knowledge, thereby preserving the structure and details of the images during the generation process[21]. We can derive a matrix $\boldsymbol{A}^{n-N}$ ($N$ is the number of max time step), thereby enabling the diffusion equation to be expressed in the following form:

$$\boldsymbol{u}_n = \sqrt{\alpha_n}\boldsymbol{A}\boldsymbol{u}_{n-1} + \sqrt{1-\alpha_n}\boldsymbol{A}^{n-N}\boldsymbol{\varepsilon}_{n-1},\ \boldsymbol{\varepsilon}_n \sim \mathcal{N}(0, \boldsymbol{I}) \tag{3}$$

where $0 < \alpha_n < 1$. Finally, we can obtain:

$$\boldsymbol{u}_n = \sqrt{\bar{\alpha}_n}\boldsymbol{A}^n\boldsymbol{u}_0 + \boldsymbol{A}^{n-N}\boldsymbol{e}'_n \tag{4}$$

where

$$\boldsymbol{e}'_n = \sqrt{\alpha_n \cdots \alpha_2(1-\alpha_1)}\boldsymbol{\varepsilon}_0 + \sqrt{\alpha_n \cdots \alpha_3(1-\alpha_2)}\boldsymbol{\varepsilon}_1 + \cdots + \sqrt{\alpha_n(1-\alpha_{n-1})}\boldsymbol{\varepsilon}_{n-2}$$

$$+\sqrt{1-\alpha_n}\boldsymbol{\varepsilon}_{n-1} \sim \mathcal{N}(0, (1-\bar{\alpha}_n)\boldsymbol{I})$$

Rearranging Eq. (4) yields:

$$\boldsymbol{u}_n = \sqrt{\bar{\alpha}_n}\boldsymbol{A}^n\boldsymbol{u}_0 + \sqrt{1-\bar{\alpha}_n}\boldsymbol{A}^{n-N}\boldsymbol{e}_n,\ \boldsymbol{e}_n \sim \mathcal{N}(0, \boldsymbol{I}) \tag{5}$$

To simplify the equation and facilitate an intuitive understanding of its variations, we have:

$$\boldsymbol{u}_N = \sqrt{\bar{\alpha}_N}\boldsymbol{A}^N\boldsymbol{u}_0 + \sqrt{1-\bar{\alpha}_N}\boldsymbol{e}_N$$

when $N$ is sufficiently large, $\bar{\alpha}_N = \alpha_N \cdots \alpha_1 \to 0$,then $\boldsymbol{u}_N \to \boldsymbol{e}_N \sim \mathcal{N}(0, \boldsymbol{I})$. Based on Eq. (3) and Eq. (5) the forward transition probability of the noise-adding process can be derived as follows:

$$q(\boldsymbol{u}_n|\boldsymbol{u}_{n-1}) = \mathcal{N}\big(\boldsymbol{u}_n|\sqrt{\alpha_n}\boldsymbol{A}\boldsymbol{u}_{n-1}, (1-\alpha_n)\boldsymbol{B}_n\big)$$

$$q(\boldsymbol{u}_n|\boldsymbol{u}_0) = \mathcal{N}\big(\boldsymbol{u}_n|\sqrt{\bar{\alpha}_n}\boldsymbol{A}^n\boldsymbol{u}_0, (1-\bar{\alpha}_n)\boldsymbol{B}_n\big)$$

where $\boldsymbol{B}_n = \boldsymbol{A}^{n-N}(\boldsymbol{A}^{n-N})^T$, which is evidently symmetric.

The backward transition probability of the diffusion process is:

$$q(\boldsymbol{u}_{n-1}|\boldsymbol{u}_n,\boldsymbol{u}_0) = \frac{q(\boldsymbol{u}_n|\boldsymbol{u}_{n-1})q(\boldsymbol{u}_{n-1}|\boldsymbol{u}_0)}{q(\boldsymbol{u}_n|\boldsymbol{u}_0)} =$$

$$\frac{N(\boldsymbol{u}_n|\sqrt{\alpha_n}\boldsymbol{A}\boldsymbol{u}_{n-1},(1-\alpha_n)\boldsymbol{B}_n)N(\boldsymbol{u}_{n-1}|\sqrt{\bar{\alpha}_{n-1}}\boldsymbol{A}^{n-1}\boldsymbol{u}_0,(1-\bar{\alpha}_{n-1})\boldsymbol{B}_{n-1})}{N(\boldsymbol{u}_n|\sqrt{\bar{\alpha}_n}\boldsymbol{A}^n\boldsymbol{u}_0,(1-\bar{\alpha}_n)\boldsymbol{B}_n)}$$

the exponential term is a quadratic form in $\boldsymbol{u}_{n-1}$, and the part involving only $\boldsymbol{u}_{n-1}$ is written as:

$$\propto exp\left\{-\frac{1}{2}\left[\left(\boldsymbol{u}_n-\sqrt{\alpha_n}\boldsymbol{A}\boldsymbol{u}_{n-1}\right)^T(1-\alpha_n)^{-1}\boldsymbol{B}_n^{-1}\left(\boldsymbol{u}_n-\sqrt{\alpha_n}\boldsymbol{A}\boldsymbol{u}_{n-1}\right)\right.\right.$$
$$+\left(\boldsymbol{u}_{n-1}-\sqrt{\bar{\alpha}_{n-1}}\boldsymbol{A}^{n-1}\boldsymbol{u}_0\right)^T(1-\bar{\alpha}_{n-1})^{-1}\boldsymbol{B}_{n-1}^{-1}\left(\boldsymbol{u}_{n-1}\right.$$
$$\left.\left.\left.-\sqrt{\bar{\alpha}_{n-1}}\boldsymbol{A}^{n-1}\boldsymbol{u}_0\right)\right]\right\}$$

From the quadratic term in $\boldsymbol{u}_{n-1}$ in this expression, the covariance matrix of the backward transition probability $q(\boldsymbol{u}_{n-1}|\boldsymbol{u}_n,\boldsymbol{u}_0)$ can be derived:

$$\boldsymbol{\Sigma}_n = [(1-\alpha_n)^{-1}\alpha_n\boldsymbol{A}^T\boldsymbol{B}_n^{-1}\boldsymbol{A} + (1-\bar{\alpha}_{n-1})^{-1}\boldsymbol{B}_{n-1}^{-1}]^{-1}$$

By setting the gradients of the quadratic and linear terms with respect to $\boldsymbol{u}_{n-1}$ to zero:

$$2(1-\alpha_n)^{-1}\alpha_n\boldsymbol{A}^T\boldsymbol{B}_n^{-1}\boldsymbol{A}\boldsymbol{u}_{n-1} + 2(1-\bar{\alpha}_{n-1})^{-1}\boldsymbol{B}_{n-1}^{-1}\boldsymbol{u}_{n-1}$$
$$-(1-\alpha_n)^{-1}\sqrt{\alpha_n}\boldsymbol{A}^T\boldsymbol{B}_n^{-1}\boldsymbol{u}_n-(1-\alpha_n)^{-1}\sqrt{\alpha_n}\boldsymbol{A}^T\boldsymbol{B}_n^{-1}\boldsymbol{u}_n$$
$$-(1-\bar{\alpha}_{n-1})^{-1}\sqrt{\bar{\alpha}_{n-1}}\boldsymbol{B}_{n-1}^{-1}\boldsymbol{A}^{n-1}\boldsymbol{u}_0$$
$$-(1-\bar{\alpha}_{n-1})^{-1}\sqrt{\bar{\alpha}_{n-1}}\boldsymbol{B}_{n-1}^{-1}\boldsymbol{A}^{n-1}\boldsymbol{u}_0 = 0$$

the mean of the backward transition probability $q(\boldsymbol{u}_{n-1}|\boldsymbol{u}_n,\boldsymbol{u}_0)$ can be obtained:

$$\boldsymbol{\mu}_n(\boldsymbol{u}_n,\boldsymbol{u}_0) = [(1-\alpha_n)^{-1}\alpha_n\boldsymbol{A}^T\boldsymbol{B}_n^{-1}\boldsymbol{A} + (1-\bar{\alpha}_{n-1})^{-1}\boldsymbol{B}_{n-1}^{-1}]^{-1}$$
$$\cdot\left[(1-\alpha_n)^{-1}\sqrt{\alpha_n}\boldsymbol{A}^T\boldsymbol{B}_n^{-1}\boldsymbol{u}_n + (1-\bar{\alpha}_{n-1})^{-1}\sqrt{\bar{\alpha}_{n-1}}\boldsymbol{B}_{n-1}^{-1}\boldsymbol{A}^{n-1}\boldsymbol{u}_0\right] \quad (6)$$

That is:

$$q(\boldsymbol{u}_{n-1}|\boldsymbol{u}_n,\boldsymbol{u}_0) = \mathcal{N}(\boldsymbol{u}_{n-1}|\boldsymbol{\mu}_n(\boldsymbol{u}_n,\boldsymbol{u}_0),\boldsymbol{\Sigma}_n) \quad (7)$$

While $n = 1$:

$$q(\boldsymbol{u}_0'|\boldsymbol{u}_1,\boldsymbol{u}_0) = \delta(\boldsymbol{u}_0' - \boldsymbol{u}_0)$$

At this point, the formulas for the diffusion process have been fully updated.

The transition probability of the generative process:

$$p_\theta(\boldsymbol{u}_{n-1}|\boldsymbol{u}_n) = \begin{cases}\mathcal{N}(\boldsymbol{u}_{n-1}|\widehat{\boldsymbol{\mu}}_\theta(\boldsymbol{u}_n),\boldsymbol{\Sigma}_n), n>1\\ \mathcal{N}(\boldsymbol{u}_0|\widehat{\boldsymbol{\mu}}_\theta(\boldsymbol{u}_1),\boldsymbol{I}), n=1\end{cases} \quad (8)$$

Rewrite

$$\boldsymbol{u}_n = \sqrt{\bar{\alpha}_n}\boldsymbol{A}^n\boldsymbol{u}_0 + \sqrt{1-\bar{\alpha}_n}\boldsymbol{A}^{n-N}\boldsymbol{e}_n$$

as

$$\boldsymbol{u}_0 = \frac{1}{\sqrt{\bar{\alpha}_n}}\left(\boldsymbol{A}^{-n}\boldsymbol{u}_n - \sqrt{1-\bar{\alpha}_n}\boldsymbol{A}^{-N}\boldsymbol{e}_n\right)$$

Substituting it into Eq. (7) yields:

$$\boldsymbol{\mu}_n(\boldsymbol{u}_n, \boldsymbol{u}_0) = \boldsymbol{C}_n\boldsymbol{u}_n + \boldsymbol{D}_n\boldsymbol{e}_n \tag{9}$$

where

$$\boldsymbol{C}_n = [(1-\alpha_n)^{-1}\alpha_n\boldsymbol{A}^T\boldsymbol{B}_n^{-1}\boldsymbol{A} + (1-\bar{\alpha}_{n-1})^{-1}\boldsymbol{B}_{n-1}^{-1}]^{-1} \cdot \left[\frac{\sqrt{\alpha_n}}{1\text{-}\alpha_n}\boldsymbol{A}^T\boldsymbol{B}_n^{-1} + \frac{1}{(1-\bar{\alpha}_{n-1})\sqrt{\alpha_n}}\boldsymbol{B}_{n-1}^{-1}\boldsymbol{A}^{-1}\right]$$

$$\boldsymbol{D}_n = -\frac{\sqrt{1-\bar{\alpha}_n}}{(1-\bar{\alpha}_{n-1})\sqrt{\alpha_n}}[(1-\alpha_n)^{-1}\alpha_n\boldsymbol{A}^T\boldsymbol{B}_n^{-1}\boldsymbol{A} + (1-\bar{\alpha}_{n-1})^{-1}\boldsymbol{B}_{n-1}^{-1}]^{-1}\boldsymbol{B}_{n-1}^{-1}\boldsymbol{A}^{n-N-1}$$

According to $\boldsymbol{\mu}_1(\boldsymbol{u}_1, \boldsymbol{u}_0) = \boldsymbol{u}_0$ and $\boldsymbol{u}_0 = \frac{1}{\sqrt{\bar{\alpha}_n}}\left(\boldsymbol{A}^{-n}\boldsymbol{u}_n - \sqrt{1-\bar{\alpha}_n}\boldsymbol{A}^{-N}\boldsymbol{e}_n\right)$, when $n-1$:

$$\boldsymbol{u}_0 = \boldsymbol{\mu}_1(\boldsymbol{u}_1, \boldsymbol{u}_0) = \boldsymbol{C}_1\boldsymbol{u}_1 + \boldsymbol{D}_1\boldsymbol{e}_1$$

where $\boldsymbol{C}_1 = \frac{1}{\sqrt{\alpha_1}}\boldsymbol{A}^{-1}$, $\boldsymbol{D}_1 = -\sqrt{\frac{1-\alpha_1}{\alpha_1}}\boldsymbol{A}^{-N}$. Following the form of Eq. (9), we can obtain:

$$\widehat{\boldsymbol{\mu}}_\theta(\boldsymbol{u}_n) = \boldsymbol{C}_n\boldsymbol{u}_n + \boldsymbol{D}_n\widehat{\boldsymbol{e}}_\theta(\boldsymbol{u}_n) \tag{10}$$

Now, we have derived the corresponding formulas for both the diffusion process and the generative process of DDPM when incorporating inter-pixel relationships.

We need to employ the ELBO to optimize the objective and identify the distribution that most closely approximates the true distribution[22].

$$\mathrm{ELBO}_\theta(\boldsymbol{u}_0) = E_{q(\boldsymbol{u}_1|\boldsymbol{u}_0)}[\ln p_\theta(\boldsymbol{u}_0|\boldsymbol{u}_1)] - \mathrm{D_{KL}}\big(q(\boldsymbol{u}_N|\boldsymbol{u}_0)||p(\boldsymbol{u}_N)\big) - \textstyle\sum_{n=2}^{N} E_{q(\boldsymbol{u}_n|\boldsymbol{u}_0)}\left[\mathrm{D_{KL}}\big(q(\boldsymbol{u}_{n-1}|\boldsymbol{u}_n, \boldsymbol{u}_0)||p_\theta(\boldsymbol{u}_{n-1}|\boldsymbol{u}_n)\big)\right]$$

where, based on Eq. (7), Eq. (8), Eq. (9) and Eq. (10), the KL divergence can be derived as follows:

$$\mathrm{D_{KL}}\big(q(\boldsymbol{u}_{n-1}|\boldsymbol{u}_n, \boldsymbol{u}_0)||p_\theta(\boldsymbol{u}_{n-1}|\boldsymbol{u}_n)\big)$$

$$= \tfrac{1}{2}[(\boldsymbol{\mu}_n(\boldsymbol{u}_n, \boldsymbol{u}_0) - \widehat{\boldsymbol{\mu}}_\theta(\boldsymbol{u}_n))^T\boldsymbol{\Sigma}_n^{-1}(\boldsymbol{\mu}_n(\boldsymbol{u}_n, \boldsymbol{u}_0) - \widehat{\boldsymbol{\mu}}_\theta(\boldsymbol{u}_n))]$$

$$= \tfrac{1}{2}\left[\big(\boldsymbol{e}_n - \widehat{\boldsymbol{e}}_\theta(\boldsymbol{u}_n)\big)^T\boldsymbol{D}_n^T\boldsymbol{\Sigma}_n^{-1}\boldsymbol{D}_n\big(\boldsymbol{e}_n - \widehat{\boldsymbol{e}}_\theta(\boldsymbol{u}_n)\big)\right]$$

thus, we have:

$$\mathrm{ELBO}_\theta(\boldsymbol{u}_0) = E_{q(\boldsymbol{u}_1|\boldsymbol{u}_0)}[\ln p_\theta(\boldsymbol{u}_0|\boldsymbol{u}_1)] - \frac{1}{2}\sum_{n=2}^{N} E_{q(\boldsymbol{u}_n|\boldsymbol{u}_0)} \left[\big(\boldsymbol{e}_n - \widehat{\boldsymbol{e}}_\theta(\boldsymbol{u}_n)\big)^T\boldsymbol{D}_n^T\boldsymbol{\Sigma}_n^{-1}\boldsymbol{D}_n\big(\boldsymbol{e}_n - \widehat{\boldsymbol{e}}_\theta(\boldsymbol{u}_n)\big)\right] + \mathrm{Const} \tag{11}$$

based on Eq. (8), Eq. (10) and $\boldsymbol{u}_0=\boldsymbol{\mu}_1(\boldsymbol{u}_1,\boldsymbol{u}_0)=\boldsymbol{C}_1\boldsymbol{u}_1+\boldsymbol{D}_1\boldsymbol{e}_1$, it can be concluded that (where $\boldsymbol{\Sigma}_1=\boldsymbol{I}$):

$$\ln p_\theta(\boldsymbol{u}_0|\boldsymbol{u}_1)=-\frac{1}{2}\left[\left(\boldsymbol{u}_0-\widehat{\boldsymbol{\mu}}_\theta(\boldsymbol{u}_1)\right)^T\boldsymbol{\Sigma}_1^{-1}\left(\boldsymbol{u}_0-\widehat{\boldsymbol{\mu}}_\theta(\boldsymbol{u}_1)\right)\right]+\text{Const}$$

$$=-\frac{1}{2}\left[\left(\boldsymbol{u}_0-\boldsymbol{C}_1\boldsymbol{u}_1-\boldsymbol{D}_1\hat{\boldsymbol{e}}_\theta(\boldsymbol{u}_1)\right)^T\boldsymbol{\Sigma}_1^{-1}\left(\boldsymbol{u}_0-\boldsymbol{C}_1\boldsymbol{u}_1-\boldsymbol{D}_1\hat{\boldsymbol{e}}_\theta(\boldsymbol{u}_1)\right)\right]+\text{Const}$$

$$=-\frac{1}{2}\left[\left(\boldsymbol{D}_1\boldsymbol{e}_1-\boldsymbol{D}_1\hat{\boldsymbol{e}}_\theta(\boldsymbol{u}_1)\right)^T\boldsymbol{\Sigma}_1^{-1}\left(\boldsymbol{D}_1\boldsymbol{e}_1-\boldsymbol{D}_1\hat{\boldsymbol{e}}_\theta(\boldsymbol{u}_1)\right)\right]+\text{Const}$$

$$=-\frac{1}{2}\left[\left(\boldsymbol{e}_1-\hat{\boldsymbol{e}}_\theta(\boldsymbol{u}_1)\right)^T\boldsymbol{D}_1^T\boldsymbol{\Sigma}_1^{-1}\boldsymbol{D}_1\left(\boldsymbol{e}_1-\hat{\boldsymbol{e}}_\theta(\boldsymbol{u}_1)\right)\right]+\text{Const}$$

from this, the final ELBO can be calculated as:

$$\text{ELBO}_\theta(\boldsymbol{u}_0)=-\frac{1}{2}\sum_{n=1}^{N}E_{q(\boldsymbol{u}_n|\boldsymbol{u}_0)}\left[\left(\boldsymbol{e}_n-\hat{\boldsymbol{e}}_\theta(\boldsymbol{u}_n)\right)^T\boldsymbol{D}_n^T\boldsymbol{\Sigma}_n^{-1}\boldsymbol{D}_n\left(\boldsymbol{e}_n-\hat{\boldsymbol{e}}_\theta(\boldsymbol{u}_n)\right)\right]+\text{Const}$$

The following two tables respectively outline the program execution sequence of the diffusion process and the generation process.

**Algorithm 1** Training

1: Receive $\boldsymbol{u}_0$ from training data
2: **while** not converged do
3: Sample $n\sim\mathcal{U}(1,N)$
4: Sample $\boldsymbol{e}_n\sim\mathcal{N}(0,\boldsymbol{I})$
5: Compute $\boldsymbol{u}_n=\sqrt{\bar{\alpha}_n}\boldsymbol{A}^n\boldsymbol{u}_0+\sqrt{1-\bar{\alpha}_n}\boldsymbol{A}^{n-N}\boldsymbol{e}_n$
6: Update $\boldsymbol{\theta}\leftarrow\boldsymbol{\theta}-\eta\nabla_{\boldsymbol{\theta}}\left[\left(\boldsymbol{e}_n-\hat{\boldsymbol{e}}_\theta(\boldsymbol{u}_n)\right)^T\boldsymbol{D}_n^T\boldsymbol{\Sigma}_n^{-1}\boldsymbol{D}_n\left(\boldsymbol{e}_n-\hat{\boldsymbol{e}}_\theta(\boldsymbol{u}_n)\right)\right]$
7: **end while**
8: return $\boldsymbol{\theta}$

**Algorithm 2** Sampling

1: Sample $\boldsymbol{u}_N\sim\mathcal{N}(0,\boldsymbol{I})$
2: **For** $n=N\sim1$
3: Sample $\boldsymbol{z}_n\sim\mathcal{N}(0,\boldsymbol{I})$
4: Cholesky decomposition $\boldsymbol{\Sigma}_n=\boldsymbol{L}_n\boldsymbol{L}_n^T$
5: Compute $\boldsymbol{u}_{n-1}=\boldsymbol{C}_n\boldsymbol{u}_n+\boldsymbol{D}_n\hat{\boldsymbol{e}}_\theta(\boldsymbol{u}_n)+\boldsymbol{L}_n\boldsymbol{z}_n$
6: **end for**
7: return $\boldsymbol{u}_0$

It is necessary to supplement some knowledge regarding the selection of the heat equation parameter $K$ as it influences the parameter values in subsequent experiments. The d-dimensional Fourier transform is given by:

$$\frac{1}{(2\pi)^{d/2}(\det\boldsymbol{\Sigma})^{1/2}}\exp\left[-\frac{1}{2}(\boldsymbol{x}-\boldsymbol{\mu})^T\Sigma^{-1}(\boldsymbol{x}-\boldsymbol{\mu})\right]\xrightarrow{\text{FT}}\exp\left[-\frac{1}{2}\boldsymbol{\omega}^T\boldsymbol{\Sigma}\boldsymbol{\omega}-j\boldsymbol{\mu}^T\boldsymbol{\omega}\right]$$

if $\boldsymbol{\Sigma}=\sigma^2\boldsymbol{I}$, then:

$$\frac{1}{(2\pi)^{d/2}\sigma^d}\exp\left(-\frac{1}{2\sigma^2}\|\boldsymbol{x}-\boldsymbol{\mu}\|^2\right)\xrightarrow{\text{FT}}\exp\left(-\frac{\sigma^2}{2}\|\boldsymbol{\omega}\|^2-j\boldsymbol{\mu}^T\boldsymbol{\omega}\right)\tag{12}$$

For the initial value problem:

$$\begin{cases} \frac{\partial U}{\partial t} - \kappa\nabla^2 u = 0 \\ u(\boldsymbol{x},0) = u_0(\boldsymbol{x}) \end{cases}$$

assuming $U(\boldsymbol{\omega},t)$ is the Fourier transform of $u(\boldsymbol{x},t)$ with respect to $x$, and $U_0(\boldsymbol{\omega})$ is the Fourier transform of $u_0(\boldsymbol{x})$, then:

$$\frac{\partial U}{\partial t} + \kappa\|\boldsymbol{\omega}\|^2 U = 0$$

its solution is:

$$U = \begin{cases} U_0\exp[-\kappa\|\boldsymbol{\omega}\|^2 t], t > 0 \\ 0, t \leq 0 \end{cases}$$

by setting $\sigma = \sqrt{2\kappa t}$, the following can be derived from Eq. (12):

$$\frac{1}{(4\pi\kappa t)^{d/2}}\exp\left(-\frac{1}{4\kappa t}\|\boldsymbol{x}\|^2\right) \xrightarrow{\text{FT}} \exp(-\kappa t\|\boldsymbol{\omega}\|^2)$$

According to the properties of the Fourier transform, the inverse Fourier transform of the product of two functions is the convolution of their respective inverse Fourier transforms:

$$u(\boldsymbol{x},t) = \begin{cases} \int G(\boldsymbol{x}-\boldsymbol{x}',t)U_0(\boldsymbol{x}')d\boldsymbol{x}', t > 0 \\ 0, t \leq 0 \end{cases}$$

where:

$$G(\boldsymbol{x},t) = \frac{1}{(4\pi\kappa t)^{d/2}}\exp\left(-\frac{1}{4\kappa t}\|\boldsymbol{x}\|^2\right)$$

By setting $U_0(\boldsymbol{x}) = \delta(\boldsymbol{x})$, we obtain $u(\boldsymbol{x},t) = G(\boldsymbol{x},t)$. By fixing $x$, the curve of $G(\boldsymbol{x},t)$ with respect to $t$ is as shown in Figure 1:

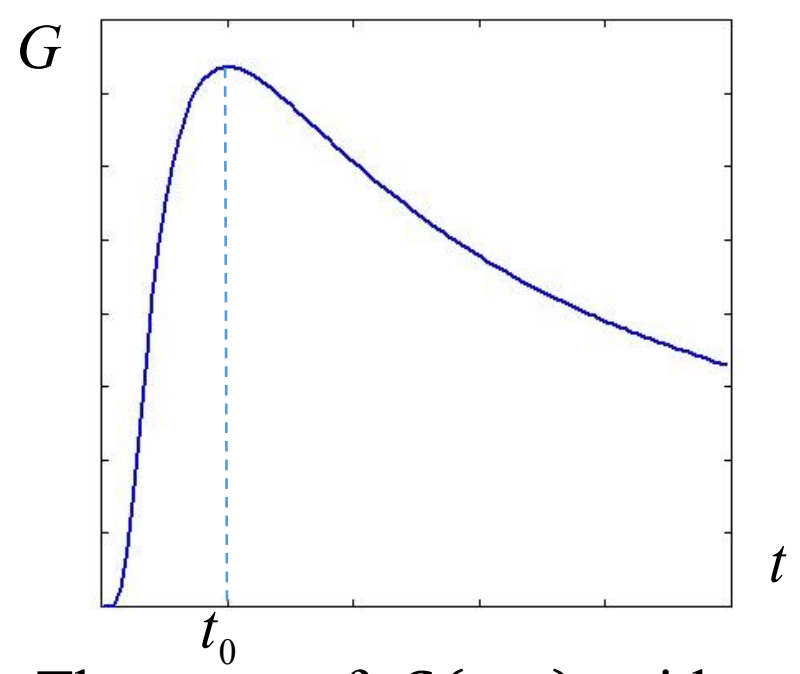

Fig. 1. The curve of $G(\boldsymbol{x},t)$ with respect to $t$

The peak time can be determined as:

$$t_0 = \frac{\|x\|^2}{2d\kappa}$$

By setting $t_0 = \tau$, $\boldsymbol{x} = (\gamma\Delta, 0, \cdots, 0)$, which implies that heat propagates $\gamma$ grid points over one time step, the above equation can be written as:

$$K = \frac{\kappa\tau}{\Delta^2} = \frac{\gamma^2}{2d}$$

When $d = 2$, $\gamma$ is chosen to satisfy $\gamma < 1/\sqrt{2}$, for example, $\gamma = 0.5$. It should be noted that the Crank-Nicolson scheme is unconditionally stable, imposing no restrictions on parameter $K$. However, matrix $A$ is invertible, requiring $K < \frac{1}{2d}$[18].

## 3. EXPERMENTAL RESULTS

### 3.1 Dataset Preparation

The dataset utilized in this experiment consists of 500 images with a resolution of 64×64 from CIFAR-10, of which 400 images were used as the training set and 100 images as the test set. From the LSUN dataset, 500 images with a resolution of 256×256 were selected from the "bridge" category. Finally, 1000 images with a resolution of 256×256 were selected from the ImageNet dataset, of which 800 images were used as the training set and 200 images as the test set.

### 3.2 Experimental Configuration

The training for this experiment was conducted on a computer equipped with 16GB of memory and an RTX 2080Ti GPU, utilizing Python version 3.9.13 and the deep learning framework PyTorch version 1.13.1.

### 3.3 Experimental Procedure and Results

In this section, we demonstrate that, under the same number of iterations as DDPM, HDM generates images of higher quality, with FID scores reduced by 0.1 points on CIFAR-10. HDM demonstrates superior performance with a 3.0 point reduction in FID on 256×256 LSUN dataset and a 4.2 point FID reduction on 256×256 ImageNet dataset. To ensure that the number of neural network evaluations required during sampling matches that of previous work, we set N=1000 for all experimen*ts.*

To determine the value of $\theta$ and $K$ in the matrix, we conducted training on 64×64 images from the CIFAR-10 dataset. Figure 2 illustrates the variation in FID scores of generated images with respect to changes in $K$ and $\theta$ values, where all FID scores were obtained with a sampling step count of 1000.

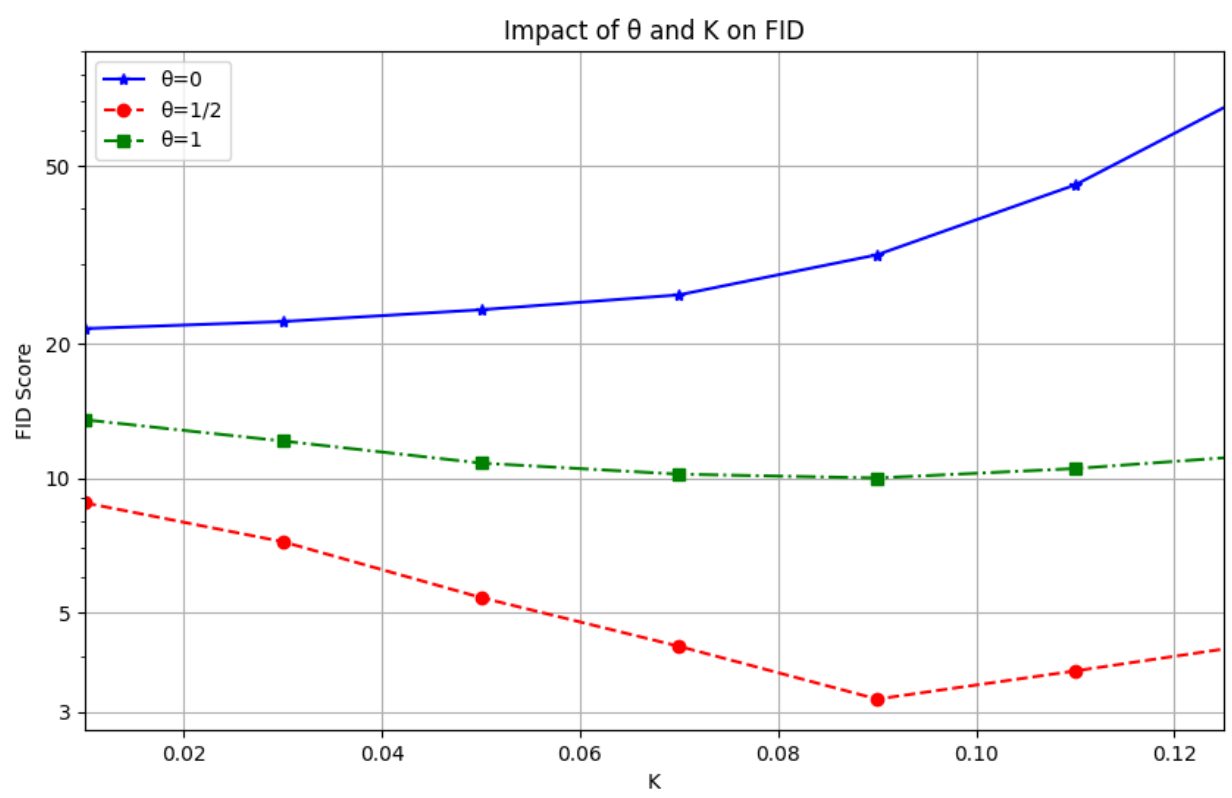

Fig. 2: The impact of $\theta$ and $K$ on the FID score.

For 0 sampling steps, the FID score is practically meaningless. At 1000 sampling steps, all three FID scores are below 50; thus, the vertical axis is simply set to 50 for zero sampling steps. The figure demonstrates that when $\theta=1/2$, the unconditionally stable scheme balances explicit efficiency with implicit accuracy, resulting in smoother interactions between adjacent pixels, enhanced detail preservation, reduced numerical oscillations, and lower FID scores for the generated images. When $\theta=1$, the pixel values of the current time step only depend on the adjacent pixel values of the previous time step. This will lead to excessive smoothing of high-frequency details or incoherent local structures, and the optimization of generation performance is limited. When $\theta=0$, the current pixel value is completely determined by the values of adjacent pixels at the current time step. This configuration suppresses high-frequency noise at the expense of fine details, resulting in a reduction in the diversity of the generated image. Therefore, $\theta$ is uniformly set to 1/2 throughout our experiment.

In matrix $\boldsymbol{A}$, another critical parameter is the coefficient $K$. Figure 3 shows the effect of the change of coefficient $K$ on the *FID* score of HDM. This image was obtained through training on the ImageNet dataset at 256×256 resolution.

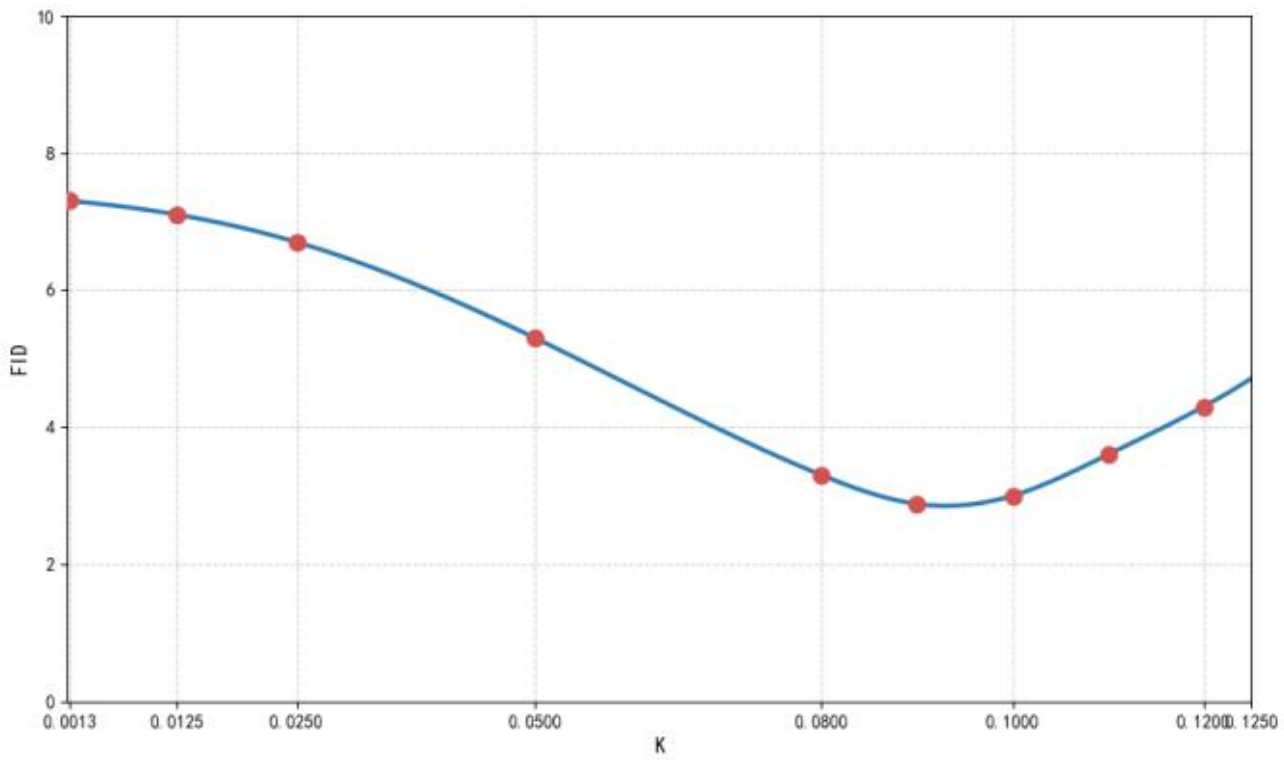

Fig. 3: The impact of the coefficient $K$ on the FID score.

When $K=0$, matrix $\boldsymbol{A}$ reduces to an identity matrix, and the corresponding FID score equals that of the standard DDPM. When $K$ is too small, the weights of the non-diagonal elements of matrix $\boldsymbol{A}$ are small, and the interactions between pixels are almost negligible. While high-frequency details can be largely preserved, the lack of inter-pixel cooperative optimization may result in isolated noise artifacts or locally incoherent regions. The optimization efficacy is not significantly pronounced. The monotonic increase in the $K$ -value initially enhances the denoising capability of the model through the introduction of structured perturbations that conform to natural image degradation patterns, thereby improving generation quality and reducing the FID score. However, as $K$ approaches the critical threshold of 0.125, the excessive attenuation of the central weighting coefficient leads to error accumulation during iterative updates, while the overly strong global dependencies between pixels induce numerical instability. This manifests as amplified high-frequency artifacts and compromised local consistency, coupled with excessive smoothing of critical high-frequency features, ultimately resulting in the observed rebound of the FID score.

Values exceeding 0.125 are meaningless due to the aforementioned stability boundary condition, rendering further discussion of FID variations in this regime unnecessary. Figure 3 demonstrates that $K$ values between 0.08 and 0.11 enable the model to achieve optimal generation performance, with marginal further improvement beyond this range. In our experiments, $K = 0.095$.

To validate that HDM also outperforms DDPM in generating high-dimensional images, additional tests were conducted on 256×256 images from the LSUN dataset and 256×256 images from the ImageNet dataset. Table 1 presents the FID scores obtained after training on the ImageNet dataset, demonstrating that HDM not only achieves better scores than DDPM but also outperforms subsequent models such as CDM and LDM. Furthermore, it outperforms the majority of models in AR.

Table 1. ImageNet 256×256 results.

| Type | Model | FID ↓ | IS ↑ | Pre ↑ | Rec ↑ | #Para | #Step |
|---|---|---|---|---|---|---|---|
| GAN | BigGAN | 6.95 | 224.5 | 0.89 | 0.38 | 112M | 1 |
| | GigaGan | 3.45 | 225.50 | 0.84 | 0.61 | 569M | 1 |
| | StyleGan-XL | 2.30 | 265.1 | 0.78 | 0.53 | 166M | 1 |
| Diff. | ADM | 10.94 | 101.00 | 0.69 | 0.63 | 554M | 250 |
| | CDM | 4.88 | 158.7 | – | – | – | 8100 |
| | LDM-4-G | 3.60 | 247.7 | – | – | 400M | 250 |
| | DiT-L/2 | 5.02 | 167.2 | 0.75 | 0.57 | 458M | 250 |
| | DiT-XL/2 | 2.27 | 278.2 | 0.83 | 0.57 | 675M | 250 |
| | **Ours** | **2.89** | 221.02 | 0.78 | 0.56 | 376M | 1000 |
| Mask | MaskGIT | 6.18 | 182.1 | 0.80 | 0.51 | 227M | 8 |
| | RCG (cond.) | 3.49 | 215.5 | – | – | 502M | 20 |
| AR | VQGAN | 15.78 | 74.3 | – | – | 1.4B | 256 |
| | VQGAN-re | 5.20 | 280.3 | – | – | 1.4B | 256 |
| | ViTVQ | 4.17 | 175.1 | – | – | 1.7B | 1024 |
| | RQTran. | 7.55 | 134.0 | – | – | 3.8B | 68 |
| | RQTran.-re | 3.80 | 323.7 | – | – | 3.8B | 68 |
| | SPAE | 4.41 | 133.03 | – | – | – | 100 |
| | LlamaGen-B | 5.46 | 193.61 | 0.83 | 0.45 | 111M | 250 |
| | LlamaGen-3B | 2.18 | 263.33 | 0.81 | 0.58 | 3B | 250 |
| VAR | VAR-d16 | 3.30 | 274.4 | 0.84 | 0.51 | 310M | 10 |
| | VAR-d20 | 2.57 | 302.6 | 0.83 | 0.56 | 600M | 10 |

Figures 4 illustrates the FID scores of HDM on the corresponding datasets from LSUN.

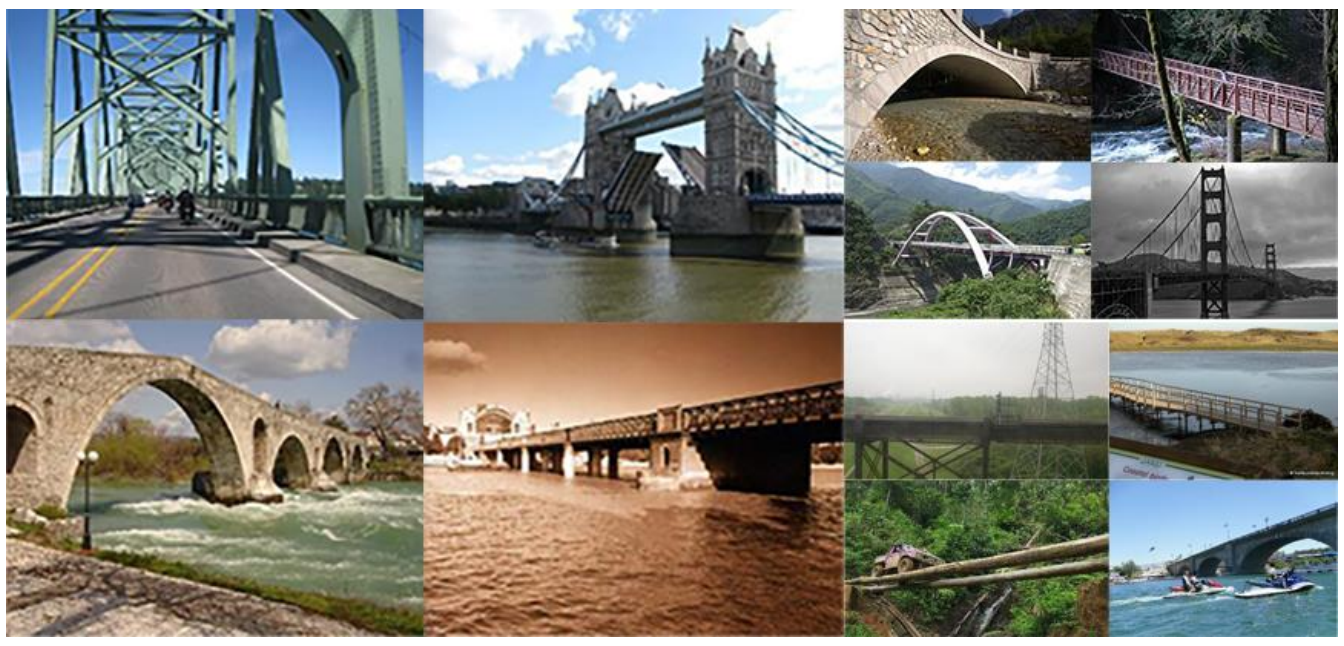

Fig. 4: LSUN Bridge samples. FID=4.22.

In order to compare the generation effect more intuitively, when we add noise to the input image in the forward process, we choose not to completely destroy the original image into a pure noise image, so as to predict the noise and achieve the denoising process.

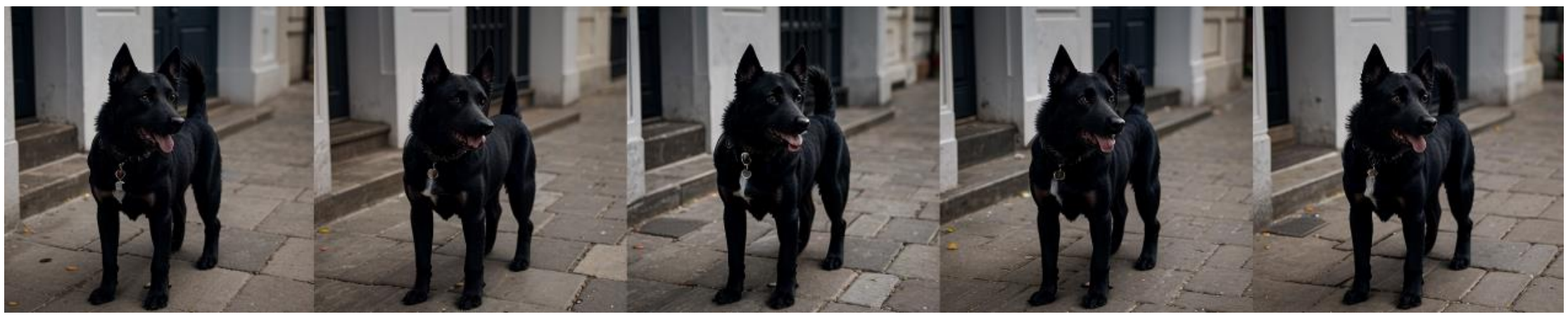

Fig. 5: Comparison: The first from the left is the original image, the second from the left is the generated image of DDPM, the very middle one is the generated image of DDIM, the second from the right is the generated image of LDM and the first from the right is the generated image of HDM.

To further validate the optimization of HDM in improving the quality of generated images, we conducted an ablation study. In the experiments, the matrix $\mathbf{A}$ representing the positional and quantitative relationships between pixels was replaced with a random matrix, where each element was assigned a randomly generated weight, while ensuring that the input and output data of the diffusion and generation processes remained in vector or matrix form. Subsequently, DDPM, HDM, and the ablation experiment were trained and tested on the ImageNet 256×256 dataset, and the FID scores of each model as a function of the number of sampling steps are illustrated in Figure 6.

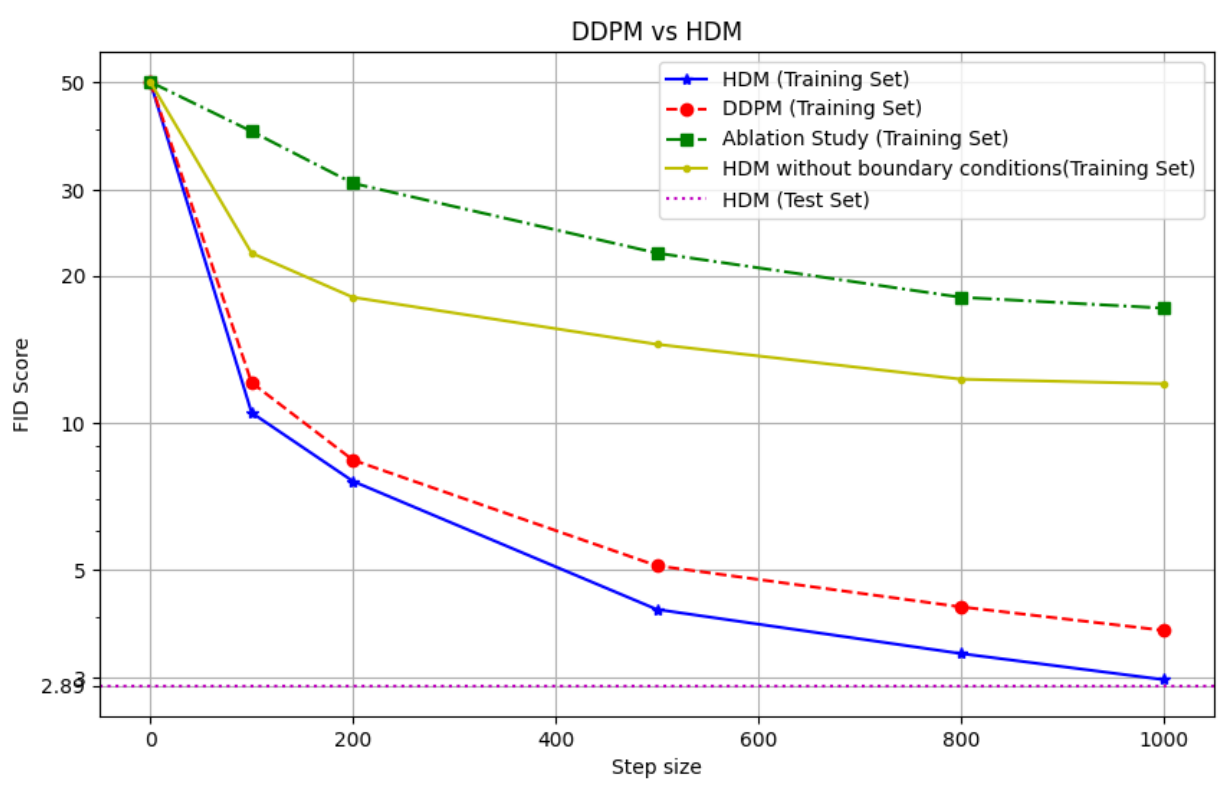

Fig. 6: The variation of FID with respect to the number of sampling steps.

As can be seen from the figure, the relational matrix incorporated in HDM achieves optimization over DDPM. The initial FID score was set to 50 for graphical convenience, and by 100 steps, the FID scores of all three models had fallen below 50.

During the experiments, we observed that the generation time for HDM and DDPM differ, however, the specific comparison of time consumption was unclear. Therefore, to determine whether the generation process of HDM requires significantly more time, we conducted a comparative experiment. The figure below illustrates the impact of the coefficient K on the generation time of HDM.

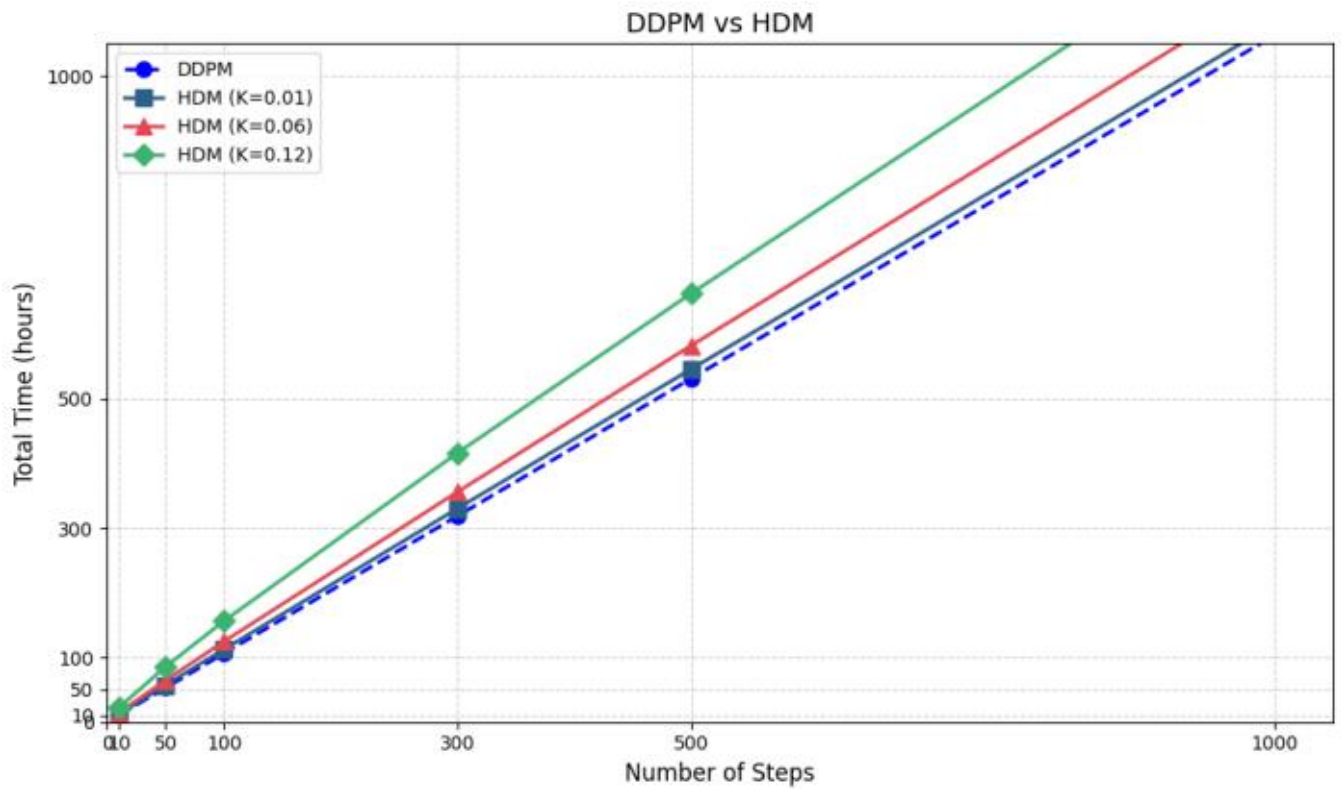

Fig. 7 Comparison of generation time among the models.

As can be observed from the figure, as the parameter K gradually increases, the time required for the model to generate images also increases significantly. This is particularly evident when the image resolution is high, as the matrix dimensionality substantially increases, causing the matrix to approach a dense matrix and leading to a notable rise in computational complexity. When the value of K is not excessively large, the time required for HDM is comparable to that of DDPM, and GPU acceleration can be utilized to expedite matrix operations. Therefore, in the experiments, to ensure image quality while optimizing performance, the optimal choice for the parameter K lies within the range of 0.08 and 0.1.

## 4. CONCLUSIONS

In response to the current extensive image processing demands, this study proposes a diffusion model optimized via the heat equation. By incorporating pixel-level relationships into both the diffusion and generation processes of DDPM, the model enhances its capability to preserve and process fine details, particularly in high-resolution image tasks. The incorporation of the discretized heat equation enables the diffusion model to process data in matrix form. The model efficiently handles large-scale matrix operations, achieving enhanced fine-grained feature processing with only marginal computational overhead. Our method demonstrates superior capability in distinguishing highly similar objects compared to DDPM, CDM, and LDM models, as experimentally verified. The image generation also exhibits finer details compared to most AR models and Mask. HDM has the potential to improve the authenticity of the generated images, which is also the work we will study next.

## ACKNOWLEDGMENTS

Thank you, Teacher Shouqing Jia, for your meticulous guidance. I couldn't have written this paper without my teacher. Thank you, my junior, for making me no longer lonely and allowing me to exchange academic ideas anytime and anywhere.